\pdfoutput=1

\documentclass[11pt]{article}

\usepackage{acl}
\usepackage{times}
\usepackage{latexsym}
\usepackage{enumitem}
\usepackage{physics}
\usepackage{amsmath}
\usepackage{tikz}
\usepackage{mathdots}
\usepackage{yhmath}
\usepackage{cancel}
\usepackage{color}
\usepackage{siunitx}
\usepackage{array}
\usepackage{multirow}
\usepackage{amssymb}
\usepackage{gensymb}
\usepackage{tabularx}
\usepackage{extarrows}
\usepackage{booktabs}
\usetikzlibrary{fadings}
\usetikzlibrary{patterns}
\usetikzlibrary{shadows.blur}
\usetikzlibrary{shapes}
\usepackage[T1]{fontenc}
\usepackage{algorithm}
\usepackage{algpseudocode}

\usetikzlibrary{shapes.geometric, arrows, positioning}
\usepackage[T1]{fontenc}

\usepackage[utf8]{inputenc}

\usepackage{microtype}

\usepackage{inconsolata}

\usepackage{xcolor}         
\usepackage{graphicx}
\usepackage{tabularx}
\usepackage{multirow,multicol}
\usepackage{tikz}
\usetikzlibrary{shapes.geometric, arrows, positioning}
\usepackage{booktabs}
\usepackage{algorithm}
\usepackage{algpseudocode}
\usepackage{setspace}
\usepackage{array}
\usepackage{makecell}
\usepackage{arydshln}
\usepackage{tikz}
\usetikzlibrary{tikzmark}
\usepackage{amsmath}
\definecolor{step1}{RGB}{226,148,108}
\definecolor{step2}{RGB}{94,148,211}
\definecolor{step3}{RGB}{110,40,107}
\definecolor{step4}{RGB}{130,162,111}
\definecolor{set3pink}{HTML}{FCCDE5}

\title{\includegraphics[height=1.4em]{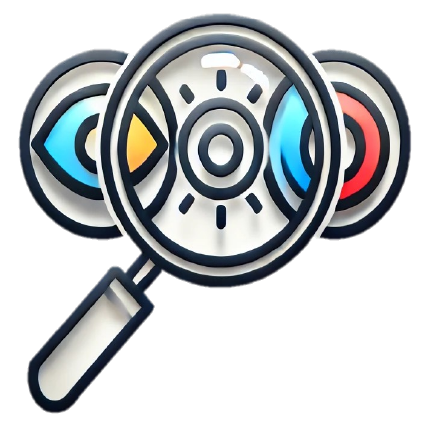}Contra4: Evaluating Contrastive Cross-Modal Reasoning in Audio, Video, Image, and 3D}

\author{Artemis Panagopoulou$^\nabla$\thanks{\tiny Work done during internship at Salesforce.} \quad
 Le Xue$^\Box$ \quad Honglu Zhou$^\Box$ \quad Silvio Savarese$^\Box$ \quad Ran Xu$^\Box$ \quad Ciaming Xiong$^\Box$ \\ \textbf{Chris Callison-Burch}$^\nabla$ \quad \textbf{Mark Yatskar}$^\nabla$ \quad \textbf{Juan Carlos Niebles}$^\Box$ \\
 $\Box$ Salesforce AI Reseach \qquad $\nabla$ University of Pennsylvania \\
  \small{Project Page:   \href{https://artemisp.github.io/contra4-web}{https://artemisp.github.io/contra4-web}}
}

\begin{document}
\maketitle
\begin{abstract}
{Real-world decision-making often begins with identifying which modality contains the most relevant information for a given query. While recent multimodal models have made impressive progress in processing diverse inputs, it remains unclear whether they can reason \textit{contrastively} across multiple modalities to select the one that best satisfies a natural language prompt. We argue this capability is foundational, especially in retrieval-augmented and decision-time contexts, where systems must evaluate multiple signals and  identify which one conveys the relevant information. To evaluate this skill, we introduce \textbf{Contra4}, a dataset for contrastive cross-modal reasoning across four modalities: image, audio, video, and 3D. Each example presents a natural language question alongside multiple candidate modality instances, and the model must select the one that semantically aligns with the prompt. Contra4 combines human-annotated captions with a mixture-of-models round-trip-consistency filter to ensure high-quality supervision, resulting in 174k training examples and a manually verified test set of 2.3k samples. While task-specific fine-tuning helps improve performance by 56\% \textit{relative} to baseline, state-of-the-art models still achieve only an \textit{absolute} of 56\% accuracy overall and 42\% in four-modality settings, underscoring a significant limitation in current multimodal models.}
\end{abstract}

\section{Introduction}
{In many real-world situations, systems are presented with multiple streams of sensory data, images, audio recordings, video clips, or 3D scans, and must determine which one best answers a specific question. For example, a security monitoring system might receive a video feed, a 3D scan, and ambient audio and be asked: “Which modality suggests someone is approaching stealthily?”. A clinician reviewing diagnostic inputs might ask: “Which of these captures evidence of respiratory distress?”, where the answer could lie in an X-ray image, a stethoscope audio clip, or a patient video. In design review, a team might ask: “Which representation best conveys the product’s motion?”, comparing a static sketch, a 3D render, and an animation. These tasks do not require combining signals, but rather understanding each modality in isolation and selecting the one that aligns best with a given prompt. This process, \textit{cross-modal semantic comparison}, is foundational to many multimodal applications, yet underexplored in current benchmarks.}

Recent advancements, such as OpenAI’s GPT-4o\footnote{\scriptsize\href{https://openai.com/index/hello-gpt-4o/}{https://openai.com/index/hello-gpt-4o/}} and Google’s Gemini~\citep{team2023gemini}, highlight the growing emphasis on models capable of comprehensively integrating diverse modalities, mirroring the multi-sensory nature of human perception. While these models promise broad multimodal capabilities, currently there is limited access to them: OpenAI's API currently offers limited access beyond image processing, and Gemini only allows for audio, video, and image inputs. Nevertheless, developing robust benchmarks remains essential to assess their performance across modalities as these features become widely available.

\begin{figure}[tb]
    \centering
\includegraphics[width=\linewidth]{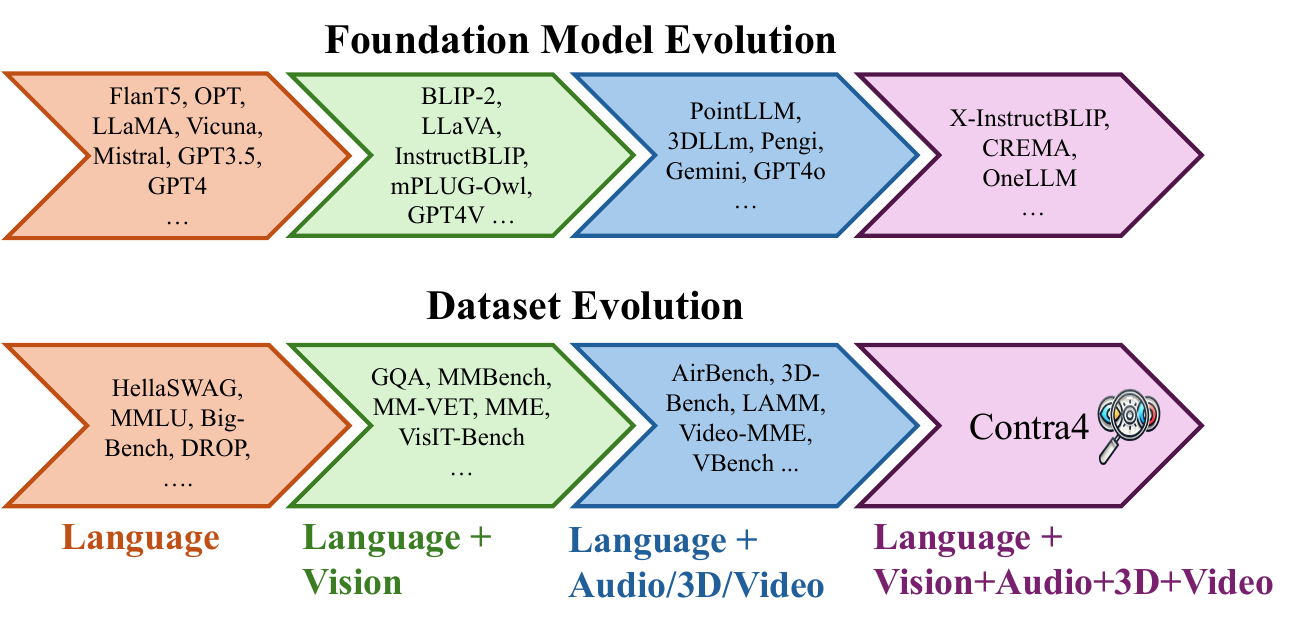}
    \vspace{-.8cm}
    \caption{\small Evolution of Foundation Models and Benchmarks. Contra4 evaluates models on multiple modalities concurrently (image, video, audio, 3D, and language).}
    \label{fig:timeline}
\end{figure}

 {Despite the growing interest in cross-modal models,\footnote{\scriptsize Cross-modal models involve 3+ modalities~\citep{panagopoulou2023x}.} there remains a significant gap in the benchmarks available to evaluate their proficiency in handling inputs across multiple modalities simultaneously. Table \ref{tab:datasets_multimodal} in the Appendix provides an overview of the major multimodal benchmarks, underscoring this deficiency. The DisCRn benchmark~\citep{panagopoulou2023x} stands out as the only dataset that integrates inputs from all four modalities. However, it lacks probing for examples that contain more than two modalities.}
 
 {To address this gap, we introduce \textbf{Contra4}, a benchmark for \textit{contrastive cross-modal reasoning} across up to four modalities: image, audio, video, and 3D. Each sample presents a natural language question alongside multiple candidate modality inputs, only one of which semantically satisfies the prompt. The dataset is intentionally designed to \textit{evaluate a critical precursor to information integration across modalities, the ability to determine which modality is most relevant to a given query}.}

{Beyond increasing the number of concurrent modalities, Contra4 includes a training set and a human-annotated test set to evaluate model performance both out-of-the-box and after fine-tuning. We implement two negative sampling strategies, high-similarity and random, to test model robustness, and apply a mixture-of-models round-trip-consistency filtering step with option permutation to ensure data quality.}\\
In summary, our contributions are the following:~\\
(i)~We introduce Contra4, a dataset requiring reasoning on up to four modalities simultaneously.~\\
(ii)~We leverage captions and a mixture-of-models round-trip-consistency strategy~(MoM-RTC) for multiple-modality data generation.~\\
(iii)~We benchmark cross-modal models and show the task's difficulty, even under fine-tuning setups.

\section{Related Work}

Advancements in vision-language tasks have paved the way for models capable of reasoning across multiple non-linguistic inputs, such as multiple images~\citep{bansal2020visual,li-etal-2022-mmcoqa,SlideVQA2023,wang-etal-2024-mementos} or cross-modal reasoning involving images and tables~\citep{li-etal-2022-mmcoqa}. Despite their complexity, these tasks predominantly focus on image-text modalities. While cross-modal benchmarks exist, primarily evaluating models on joint audio-video reasoning~\citep{alamri2018audio,li2022learning}, there remains a gap in assessing models' capabilities for comparative cross-modal reasoning. Even comprehensive multimodal benchmarks like MultiBench~\citep{liang2021multibench} and OmniXR~\citep{chen2025omnixr} primarily operate with single-modality inputs or, at most, video with corresponding audio. The medical domain follows a similar trend; for instance, M3~\citep{huang2021a} evaluates models using only corresponding X-ray images, audio, and textual input.
To address this gap, we introduce \texttt{Contra4}, a dataset to evaluate contrastive reasoning by differentiating between cross-modal inputs.

 The rise of high-performing LLMs has enabled automated data annotation, with most datasets relying on huge proprietary models like GPT-4~\citep{openai2023gpt4}. Initially used for text~\citep{dai2023auggpt,he2023annollm}, these methods now extend to images~\citep{changpinyo2021conceptual,bitton2024visit,xue2024xgen}, video~\citep{Maaz2023VideoChatGPT}, audio~\citep{wavcaps,yang2024air}, and 3D~\citep{wu20153d,zhang20243dbench}, leveraging multimodal models for alignment and LLMs for annotation. Our work differs by focusing on synthetic datasets for tasks involving 3+ modalities, where we show cross-modal reasoning remains a challenge. 
\begin{figure}[t]
    \centering
    \includegraphics[width=\linewidth]{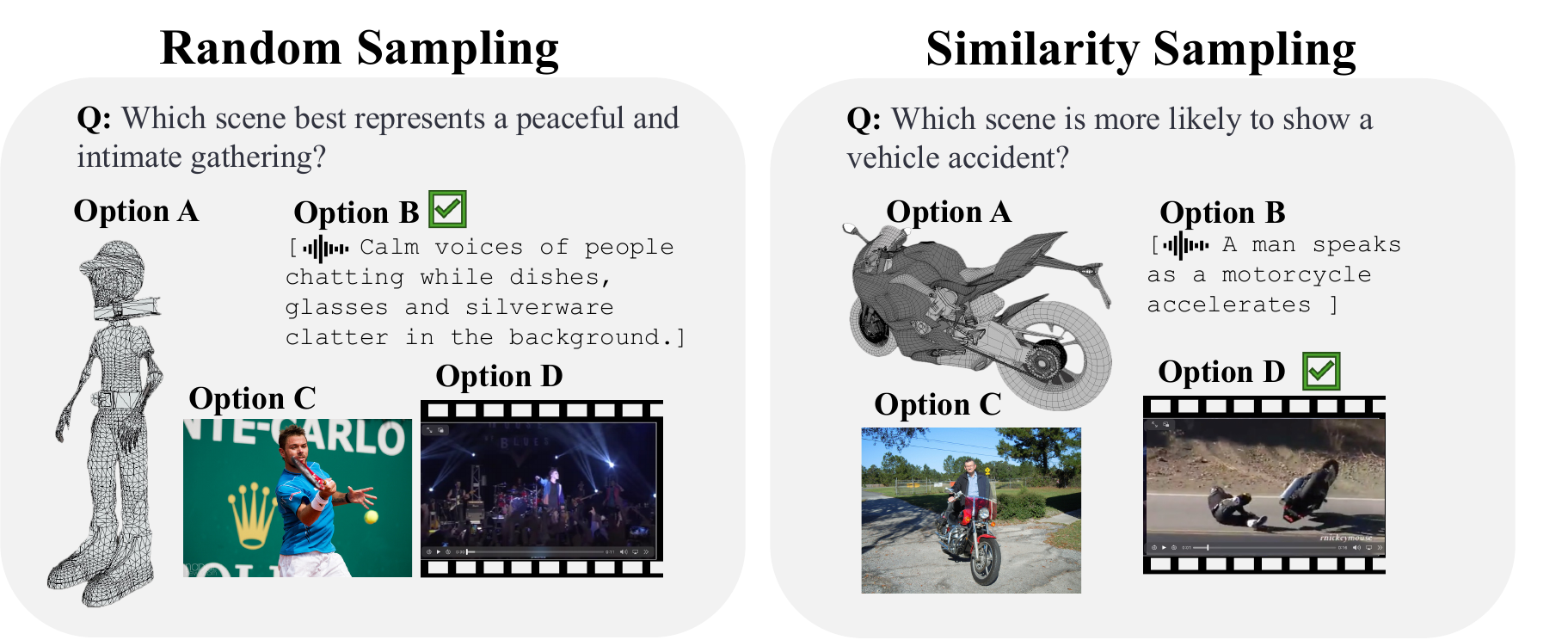}
    \vspace{-.4cm}
    \caption{\small Examples from Contra4. Additional examples are found in Figure \ref{app:qualitative_examples} in the Appendix.}
    \label{fig:examples}
\end{figure}

\section{Contra4: Task Definition}

Let $\mathbf{x} = \{x^i_{M}\}_{i=1}^N$ be a set of $N$ multimodal inputs, where each $x^i_{M}$ is drawn from a specific modality $M$ and is paired with a text query $q$, as shown in Figure \ref{fig:examples}. The function $T(\cdot)$ is used for tokenizing and embedding any textual elements, while $P_M(\cdot)$ projects an input from the modality $M$ into the model’s linguistic embedding space. In addition, each input $x^i_{M}$, encoded with a modality-specific encoder $Enc(\cdot)$, has an associated enumeration prefix $E_i$. To form the final input to the MLLM, we concatenate the tokenized prefix $T(E_i)$ with the projected multimodal representation $P_M(Enc(x^i_{M}))$ for all $i = 1, \dots, N$, and then further concatenate the tokenized query $T(q)$. Symbolically, this can be written as: {\footnotesize$
MLLM(\mathbf{x}, q) = MLLM\Bigl(\bigoplus_{i=1}^N \bigl[\,
T(E_i)\,\oplus\,P_M\bigl(Enc(x^i_{M})\bigr)\bigr] \;\oplus\; T(q)\Bigr)$},
where $\oplus$ denotes the concatenation operation in the embedding space. The model’s task is to correctly identify which enumeration prefix $E_i$ corresponds to the correct answer for the query $q$.

\begin{figure*}[ht]
    \centering
    \includegraphics[width=\linewidth]{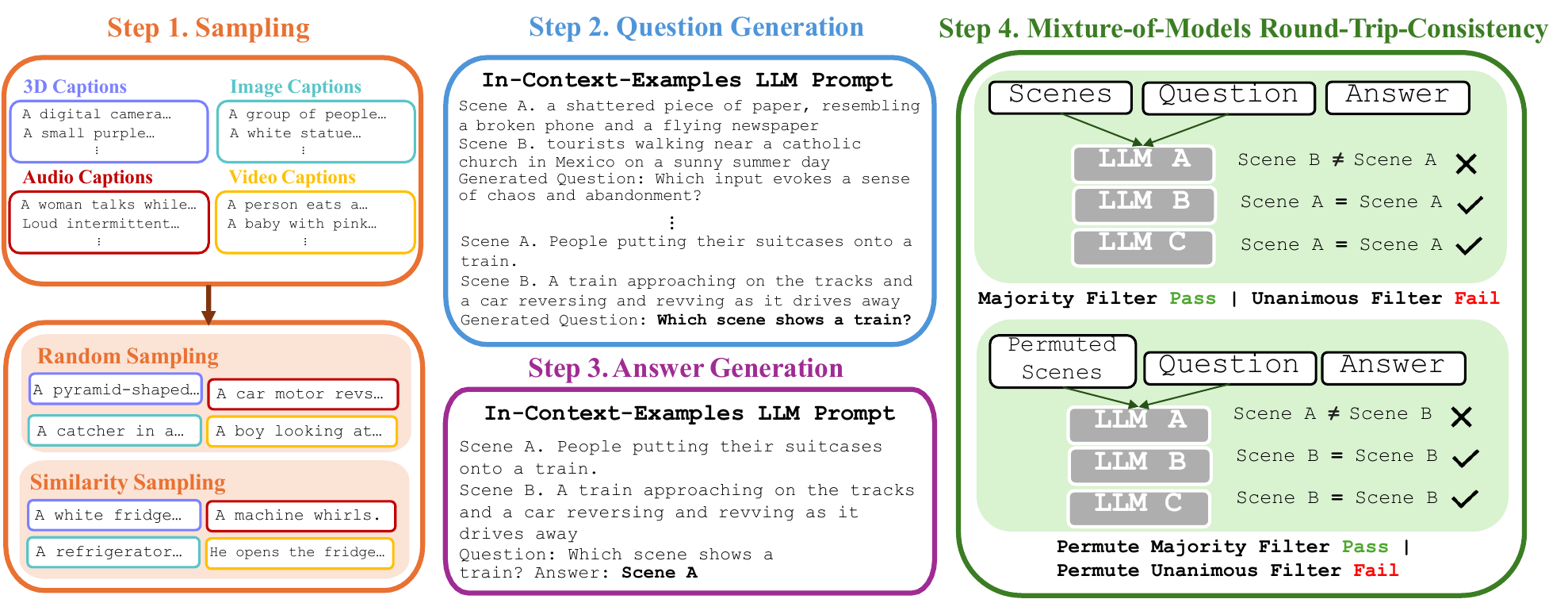}
    \vspace{-.8cm}
    \caption{\small Data Generation Pipeline. In \textcolor{step1}{Step 1}, candidate choices are sampled either randomly or by selecting those with high text similarity. \textcolor{step2}{Step 2} employs in-context learning to generate a question based on the captions, which is answered in \textcolor{step3}{Step 3}. \textcolor{step4}{Step 4} utilizes a mixture-of-models round-trip-consistency (MoM-RTC) check to eliminate incorrect samples.}
    \label{fig:overview}
    \vspace{-.4cm}
\end{figure*}

\section{Dataset}
\label{sec:data_gen}

\noindent\textbf{Data generation:} Our method leverages textual descriptions as a \textit{universal connector} across modalities to build a dataset that enables querying across diverse modalities \emph{without} requiring an additional multimodal linking model. Figure~\ref{fig:overview} illustrates our process: given a set of single-modality \(M\) datasets with associated captions, \(D_M = \{(x_M, c_M)\}\), we apply a {four-stage} data augmentation method to generate contrastive cross-modal reasoning data. \\
\noindent\textcolor{step1}{\textbf{Step 1. Negative Sampling Selection:}} We employ two negative selection strategies: \textit{high [caption] similarity} and \textit{random} to enhance the evaluation potential of the dataset. This process results in tuples of two, three, and four modalities denoted as $D_{\hat{M}}$, where $\hat{M}$ denotes the subselected modalities.\\ 
\noindent\textcolor{step2}{\textbf{Step 2. Question Generation:}} After generating tuples in \textcolor{step1}{Step 1}, we use an LLM with four in-context examples to generate a contrastive question about the multimodal inputs. Questions focusing on textual qualities of the captions are filtered out, ensuring relevance to the multimodal scene depiction.\\ \noindent\textcolor{step3}{\textbf{Step 3. Answer-Explanation Generation:}} Conditioned on the captions in the original dataset and the questions refined in \textcolor{step2}{Step 2}, we prompt the same LLM to answer and explain its reasoning. \\
\noindent\textcolor{step4}{\textbf{Step~4.~Mixture-of-Models Round-Trip-Consis-tency (MoM-RTC):}} We validate dataset quality by running a round-trip-consistency check on an ensemble of distinct models, prompting each LLM to answer and explain the contrastive questions based on their captions. We keep only samples that pass certain filtering criteria, \textbf{Majority Filter (MF)}, \textbf{Unanimous Filter (UF)}, \textbf{Permute Majority Filter (PMF)}, and \textbf{Permute Unanimous Filter (PUF)}, which we compare in Table~\ref{tab:human_eval}. In particular, MF requires that a majority of models agree with the original answer; UF requires unanimous agreement; PMF extends MF and PUF extends UF under all permutations of the cross-modal options. {Pseudocode for the procedure is presented in Algorithm \ref{alg:mom_rtc} in the Appendix for clarity.}\\
\noindent\textbf{Dataset Statistics:}
Using the above pipeline, we produce 174k automatically annotated samples for training and release a test set of 2.3k manually-annotated examples. Answer distribution is balanced post-hoc. See details in Appendix \ref{app:data_stat}.

\begin{table}[tb]
\centering
\fontsize{6}{6}\selectfont
\renewcommand{\arraystretch}{1}
\begin{tabularx}{\linewidth}{l *{7}{X}}
\toprule
Filter & Human Acc. & N/A & O/A & GPU~(hrs) & \multicolumn{3}{c}{Aggregated} \\
       &                  &     &     &           & Rand & Sim & All \\
\midrule
None$^\ast$ & 46.7  & 18.3 & 18.3 & 0 
            & 254k & 261k & 515k \\
\midrule
MF  & 60.0& 18.3 & 17.5 & \multirow{2}{*}{40}
    & 190k & 188k & 378k \\
UF  & 60.0  & 16.7 & 13.3 &
    & 130k & 126k & 256k \\
\midrule
PMF & 68.3  & 13.3& 15.0 & \multirow{2}{*}{120}
    & 147k & 131k & 278k \\
PUF & \textbf{83.3}  & \textbf{6.7} & \textbf{5.8} &
    & 91k & 83k & 174k \\
\bottomrule
\end{tabularx}
\vspace{-.2cm}
\caption{\small 
Human inspection of Round-Trip-Consistency checks on training data. 
N/A is the fraction of questions not applicable to any choice and O/A to more than one choice.
{\scriptsize $^\ast$Some rule-based word filtering is applied; see Appendix~\ref{app:data_gen}.}
}
    \label{tab:human_eval}
\end{table}

\section{Experiments}
\label{sec:experiments}

\noindent\textbf{Implementation Details:} For \textcolor{step2}{Step 2} and \textcolor{step3}{Step 3} we employ LLaMA-3.1-8B-Instruct~\citep{dubey2024llama} . For \textcolor{step4}{Step 4} we also use mistralai/Mistral-7B-Instruct-v0.2~\citep{jiang2023mistral}, and microsoft/Phi-3-medium-128k-instruct~\citep{abdin2024phi}. For the permutation checks we consider all possible permutations of the answer choices. For the text similarity we use all-MiniLM-L6-v2 encodings via \href{https://sbert.net/}{\texttt{sentence-transformers}}. Single run accuracy is reported. The datasets used to generate \texttt{Contra4} are summarized in Table \ref{tab:caption_datasets} in the Appendix with additional implementation details in Appendix \ref{app:implementation_details}.\\
\noindent\textbf{Models:} To assess task difficulty and position this dataset as a community challenge, we evaluate several state-of-the-art (SOTA) models capable of handling all four modalities. Two models, \textbf{X-InstructBLIP}~\citep{panagopoulou2023x} and \textbf{CREMA}~\citep{yu2024crema}, use a frozen LLM with separate modality encoders. They differ in that CREMA uses a fused Q-Former for modality alignment, requiring additional RGB input for 3D, whereas X-InstructBLIP maintains separate Q-Formers. We also evaluate \textbf{OneLLM}~\citep{han2023onellm}, which unifies modalities into a common space, connecting a fused modality encoder to the LLM, and trains the entire architecture, including the LLM. We also report performance of OneLLM finetuned on subsets sampled from each filtering pool in \textcolor{step4}{Step 4}. We chose OneLLM for fine-tuning because it is the top-performing model that natively processes 3D point clouds, allowing us to measure the direct impact of finetuning on the core cross-modal reasoning task without confounding factors from input signal conversion. We further baseline Gemini using \texttt{gemini-2.0-flash-exp} on examples that do not contain 3D since it is not supported. Lastly, our \textbf{Caption Baseline} replaces multimodal scenes with predicted captions for an LLM-only approach (details in Appendix~\ref{app:caption_baseline}).

\section{Discussion}

\begin{table}[tb]
    \centering
 \fontsize{5.5}{5.5}\selectfont
    \renewcommand{\arraystretch}{1}
    \begin{tabularx}{\linewidth}{p{1.3cm}l *{11}{X}}
        \toprule
        & \multicolumn{3}{c}{2 Modalities} & \multicolumn{3}{c}{3 Modalities} & \multicolumn{3}{c}{4 Modalities} & \multicolumn{3}{c}{All} \\
        \cmidrule(lr){2-4} \cmidrule(lr){5-7} \cmidrule(lr){8-10} \cmidrule(lr){11-13}
        Model & Rand. & Sim. & All &Rand. & Sim. & All & Rand. & Sim. & All & Rand. & Sim. & All \\
        \midrule
        CREMA$^{\ast}$ & \textbf{0.71} & \textbf{0.64}& \textbf{0.68} & \textbf{0.61} & \textbf{0.55} & \textbf{0.58} & \textbf{0.45} & \textbf{0.39} & \textbf{0.42}& \textbf{0.60} & \textbf{0.53} & \textbf{0.56} \\
        X-InstructBLIP & 0.47 & 0.48 & 0.47 & 0.30 & 0.27 & 0.29 & 0.13 & 0.22 & 0.18 & 0.31 & 0.33 & 0.32 \\
        OneLLM & \underline{0.52} & 0.52 & \underline{0.52} & 0.16 & 0.22 & 0.19 & 0.24 & \underline{0.27} & 0.25 & 0.31 & 0.34 & 0.32 \\
        Gemini-2.0$^\dagger$  & 0.24& 0.21&0.23&0.10&0.14 &0.13&$\times$&$\times$&$\times$&0.23&0.20&0.22 \\
        Caption Baseline & \underline{0.52} & 0.46 & 0.49 & \underline{0.33} & \underline{0.33} & \underline{0.33} & \underline{0.26} & \underline{0.27} &\underline{0.26} & \underline{0.38} & \underline{0.36} & \underline{0.37} \\
        \bottomrule
    \end{tabularx}
    \vspace{-.2cm}
        \caption{\small Zero-Shot Evaluations on Contra4 Test Set.\\ {\scriptsize  $^{\dagger}$ Proprietary LLM. Samples with 3D are excluded due to incompatibility. \\$^{\ast}$ RGB rendering signal used for 3D point clouds. }}
    \label{tab:baselines}
\end{table}

\begin{figure}[tb]
    \centering
    \includegraphics[width=\linewidth]{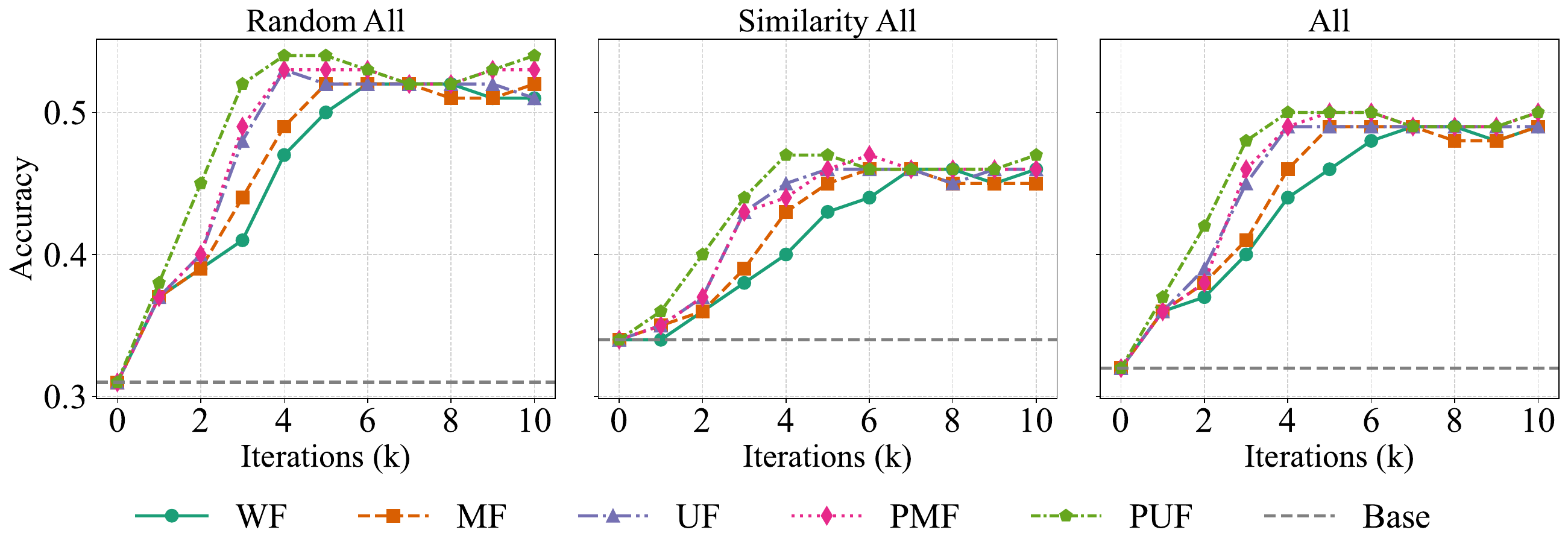}
    \vspace{-.9cm}
    \caption{\small Finetune OneLLM on different MoM-RTC data.}
    \label{fig:onellmft}
\end{figure}
\textbf{How does MoM-RTC  affect dataset quality?} We conduct a human inspection of 120 randomly selected dataset samples, evenly split by negative sampling (high-similarity vs. random) and input modality choices, to validate dataset quality and our MoM-RTC procedure. Table \ref{tab:human_eval} presents these results using the interface in Figure \ref{fig:annotation_ui}. PUF though highly selective, produces superior quality samples without relying on costly, closed-source APIs, mitigating selection bias~\citep{pezeshkpour2023large,balepur2024artifacts,wang2024my}. By admitting only examples that remain correct under choice permutations, we counteract LLM biases, improving overall correctness. 
While permutation-based RTC methods require three times the GPU hours of non-permutation approaches, they improve human-perceived precision by over 20 points.

\begin{figure}[tb]
    \centering
    \includegraphics[width=\linewidth]{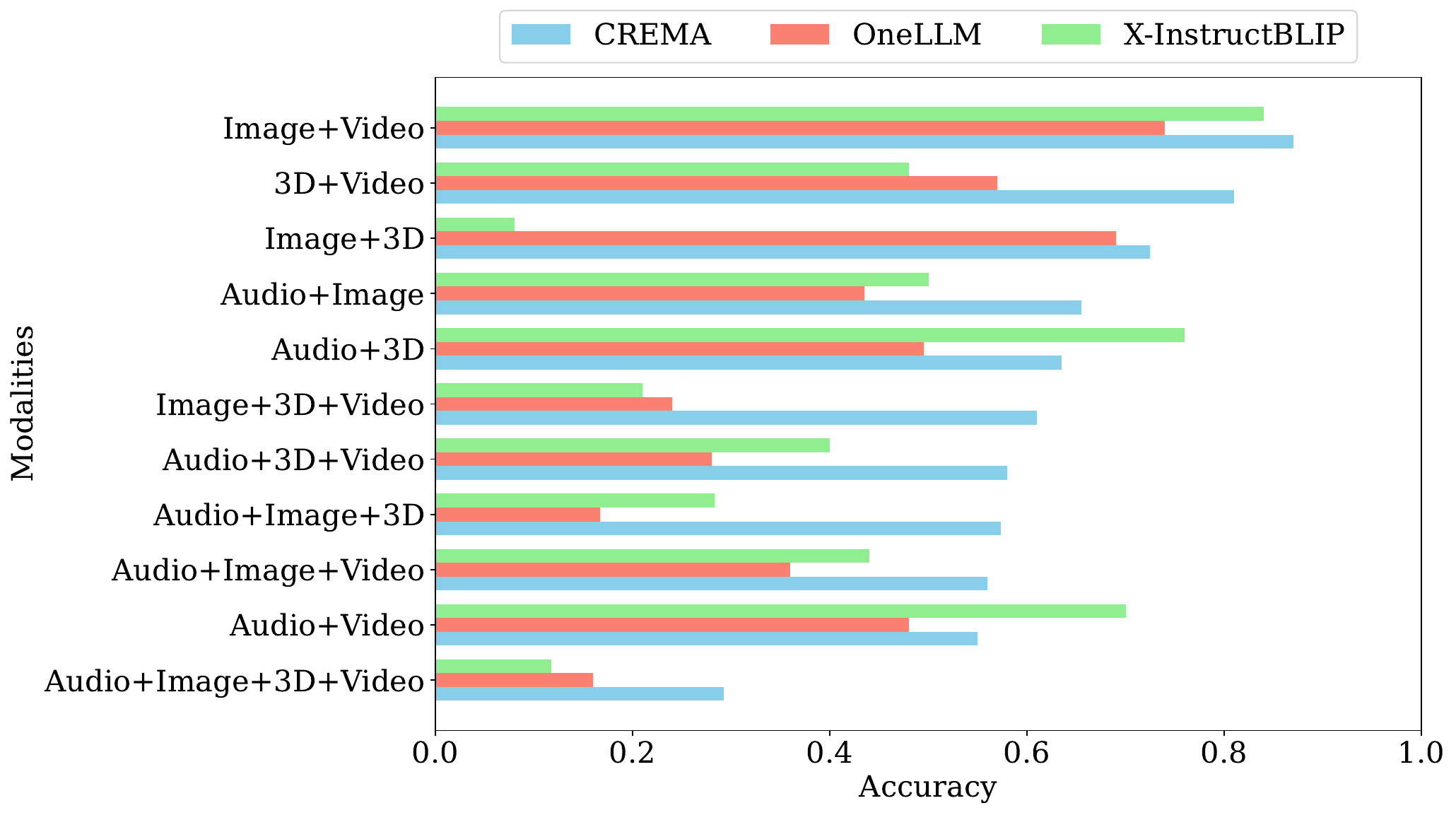}
    \vspace{-.9cm}
    \caption{\small Performance breakdown by  input modalities.}
    \label{fig:modality_breakdown}

\end{figure}

\noindent\textbf{How do SOTA models perform on Contra4?} 
Table \ref{tab:baselines} reports performance of SOTA models on the task, showing that caption-based baselines outperform most approaches as the number of modalities increases. The top performer, CREMA, relies on external RGB rendering for point clouds, though resource-intensive, it significantly boosts performance across 3D, Image, and Video (Figure \ref{fig:modality_breakdown}). While the test set is manually curated, we provided a sample of 50 examples from the test set to two independent annotators who yielded an average performance of 92\%, making CREMA's 56\% accuracy a significant gap. Architecturally, CREMA employs distinct modules for cross-modal token extraction, similar to X-InstructBLIP, but fuses them before aligning with the base LLM, aiming at a more uniform modality representation. OneLLM, in contrast, uses a fused X-modal token extraction module that appears less effective. Surprisingly, Gemini achieves the lowest score, likely due to lack of fine-tuning for this task; in many responses, it fails to recognize all three inputs, and instead resolves to captioning only the last input provided, ignoring the question. Overall, these findings suggest that systems that fuse modalities after independent extraction, followed by LLM fine-tuning, are most effective for the task. \\
\noindent\textbf{How does fine-tuning affect task performance?} To further validate our findings, we fine-tuned OneLLM on MoM-RTC data, resulting in a noticeable performance boost from 32\% to 50\%. However, overall accuracy remained low, indicating that fine-tuning alone is insufficient and alternative approaches are needed. Interestingly, despite lower human-perceived quality, all data filtering methods ultimately achieve similar performance given enough training iterations. Notably, PUF converges with the least data, followed by PMF and UF, aligning with their human-inspected accuracy rankings.\\

\indent Overall, our results reveal that even state-of-the-art multimodal models struggle with the task introduced in this work, underscoring the need for more robust cross-modal understanding.

\section{Limitations}
{A key limitation of our work lies in the simplified nature of the task formulation. While real-world applications often require models to determine which modality best supports a decision or query, Contra4 reduces this to a controlled, contrastive selection problem. \textbf{This abstraction is intentional}: it allows us to isolate a core cognitive capability, identifying the most relevant modality for a given prompt, without the confounds of task-specific logic or procedural complexity. However, we acknowledge that this setup does not fully capture the rich, dynamic nature of real-world multimodal reasoning.}

{Additionally, Contra4 emphasizes commonsense semantic matching (e.g., identifying peaceful scenes or emotional tones) rather than testing deeper inferential or task-specific reasoning.  Our goal is not to exhaustively probe all aspects of multimodal intelligence, but to provide a focused diagnostic benchmark for a foundational yet underexplored skill. That said, future work should explore benchmarks that integrate higher-level decision-making and temporal inference across modalities, and complement evaluations like Contra4 with information synthesis settings to push the boundaries of model understanding.}

Lastly, while our dataset provides a rigorous benchmark for cross-modal reasoning, performance evaluations depend on current state-of-the-art models, which may not yet be fully optimized for this task. As multimodal architectures evolve, future benchmarks should adapt accordingly to reflect their growing capabilities.

\section{Ethics Statement}

In conducting this research, we acknowledge the significant limitations and potential dangers associated with the use of Large Language Models (LLMs). One of the primary concerns is the presence of inherent biases within LLMs, which are a direct consequence of the data on which they are trained. These biases can inadvertently perpetuate harmful stereotypes and lead to discriminatory outcomes, particularly in sensitive applications. Additionally, LLMs, especially those with large parameter counts, may generate outputs that are factually incorrect or misleading, posing a risk in contexts that demand high levels of accuracy and reliability. To mitigate these risks we inspected the test samples of the dataset and used multimodal sources that would limit the potential of generation of such harmful questions. However, we emphasize the importance of ongoing vigilance and the need for responsible use of these models and our dataset to prevent unintended negative consequences.~\\

\small{\noindent\textbf{Note on AI Assistants:} AI assistants were used for grammar checks and sentence level rephrasing to improve paper flow. Coding assistants were also used to streamline development.}
\section*{Acknowledgments}
We are grateful to the reviewers for their thorough review and insightful feedback on our manuscript. This research was developed in part with funding from the Defense Advanced Research Projects Agency's (DARPA) SciFy program (Agreement No. HR00112520300). The views expressed are those of the author and do not reflect the official policy or position of the Department of Defense or the U.S. Government.

\bibliography{custom}

\appendix

\section{Data Format}
\label{app:data_format}

The dataset is stored in an easy-to-use \texttt{json} format. Each entry in the dataset consists of various fields including a unique identifier, selection type, question type, examples from various modalities, and the associated question and answer. 

\subsection{Structure}
\begin{itemize}
    \item \textbf{id}: A unique identifier for the dataset entry.
    \item \textbf{selection\_type}: The method used for selecting negative examples.
    \item \textbf{q\_type}: The question type indicating the number of choices.
    \item \textbf{examples}: A list of examples, each containing:
    \begin{itemize}
        \item \textbf{source}: The dataset from which the example is taken.
        \item \textbf{id}: A unique identifier for the example within its source.
        \item \textbf{caption}: A description of the content or scene depicted in the example.
    \end{itemize}
    \item \textbf{modalities}: A list of modalities corresponding to each example.
    \item \textbf{questions}: The question presented to the model.
    \item \textbf{answers}: The correct answer or ground truth.
    \item \textbf{category}: The category of the question, used for organizing the dataset.
\end{itemize}

\section{Benchmark Comparisons}
Table \ref{tab:datasets_multimodal} provides a succinct comparison across multimodal benchmarks,\footnote{~We do not include vision benchmarks such as GQA~\citep{hudson2019gqa}, VizWiz~\citep{bigham2010vizwiz}, and NoCaps~\citep{agrawal2019nocaps} since they appear as subsets of other benchmarks included in the table such as LVLM-eHUB~\citep{xu2023lvlm}.} showing that \texttt{Contra4} is unique in its incorporation of up to four distinct modalities in a single example. 

\begin{table*}[h]
\fontsize{7}{7}\selectfont
\centering
\begin{tabularx}{.7\textwidth}{cccccc}
\toprule
\textbf{Dataset} & \textbf{Image} & \textbf{Audio} & \textbf{Video} & \textbf{3D} & \textbf{Max Modalities per Sample} \\
\midrule
DROP~\citep{dua2019drop}              & $\times$   & $\times$   & $\times$   & $\times$   & 1 \\
MMLU~\citep{hendrycks2020measuring}   & $\times$   & $\times$   & $\times$   & $\times$   & 1 \\
MULTIBench~\citep{liang2021multibench} &  \checkmark   & \checkmark   &\checkmark   & $\times$   & 2 \\
BigBench~\citep{srivastava2023beyond} & $\times$   & $\times$   & $\times$   & $\times$   & 1\\
LVLM-eHUB~\citep{xu2023lvlm}          & \checkmark & $\times$   & $\times$   & $\times$   & 1 \\
SEED (v1)~\citep{li2023seed}          & \checkmark & $\times$   & \checkmark & $\times$   & 1 \\
SEED (v2)~\citep{li2023seed2}         & \checkmark & $\times$   & \checkmark & $\times$   & 2     \\
MM-BENCH~\citep{liu2023mmbench}        & \checkmark & $\times$   & $\times$   & $\times$   & 1 \\
VisIT-Bench~\citep{bitton2024visit}    & \checkmark & $\times$   & $\times$   & $\times$   &1 \\
MM-VET~\citep{yu2023mm}               & \checkmark & $\times$   & $\times$   & $\times$   & 1 \\
MMMU~\citep{yue2023mmmu}              & \checkmark & $\times$   & $\times$   & $\times$   & 1      \\
LAMM~\citep{yin2024lamm}              & \checkmark & $\times$   & $\times$   & \checkmark & 1 \\
AV-Superb~\citep{tseng2024av}         & \checkmark & \checkmark & \checkmark & $\times$   & 1 \\
HEAR~\citep{turian2022hear}           & $\times$   & \checkmark & $\times$   & $\times$   & 1\\
Dynamic Superb~\citep{tseng2024av}    & $\times$   & \checkmark & $\times$   & $\times$   & 1 \\
AIR-Bench~\citep{yang2024air}         & $\times$   & \checkmark & $\times$   & $\times$   & 1 \\
Video-Bench~\citep{ning2023video}     & $\times$   & \checkmark & \checkmark & $\times$   & 2 \\
3D-Bench~\citep{zhang20243dbench}     & $\times$   & $\times$   & $\times$   & \checkmark & 1 \\
OmniXR~\citep{chen2025omnixr}  & \checkmark   & \checkmark    & \checkmark    & $\times$  & 1 \\

DisCRn~\citep{panagopoulou2023x}                            & \checkmark & \checkmark & \checkmark & \checkmark & 2    \\\midrule
Contra4                                 & \checkmark & \checkmark & \checkmark & \checkmark & 4   \\
\bottomrule
\end{tabularx}
\caption{\small
\textbf{Comparison of Multimodal Challenge Datasets.} 
Columns \textit{Image}, \textit{Audio}, \textit{Video}, and \textit{3D} specify whether the dataset includes that modality (\checkmark) or not ($\times$). 
}
\label{tab:datasets_multimodal}
\end{table*}

\section{Data Generation Details}
\label{app:data_gen}
\noindent\textcolor{step1}{\textbf{Step 1: Negative Sampling Selection}} {We employ two negative selection strategies: \textit{random} and \textit{high similarity} to reduce the likelihood that the task can be solved using  unimodal heuristics.} For the high similarity negative samples, we first encode all captions across all modalities using \texttt{all-MiniLM-L6-v2} embeddings via \href{https://www.sbert.net}\texttt{sentence-transformers}. Subsequently, we anchor one modality randomly as the basis for selection. From this anchored modality, we identify and select a negative sample from among the thirty most similar instances across the different modalities, as ranked by the cosine similarity of their text captions. For the random setup, we perform the same procedure but sample randomly instead.\\
\noindent\textbf{\textcolor{step2}{Step 2 Question Generation:}} Upon generating tuples in \textcolor{step1}{Step 1}, we employ \texttt{meta-llama/Llama-3.1-8B-Instruct} to generate contrastive questions. For each tuple, we provide the LLM with four in-context examples to facilitate the generation of a question which is then considered for inclusion in the final dataset. The prompt for question generation is the following:
\begin{quote}
\small
    <s>You are given some scenes described in text. Each scene is represented by a short caption. Your task is to generate a question that compares the scenes based on their content. The generated question should be relevant to the context of the scenes and should require a comparison between them. There should be only one correct answer. Here are some examples to guide you:\\

    Scene A. "a shattered piece of paper, resembling a broken phone and a flying newspaper"\\
    Scene B. "tourists walking near a catholic church in Mexico on a sunny summer day"\\
    Generated Question: Which scene evokes a sense of chaos and abandonment?\\

    Scene A. "Someone is using a rip saw in a carpenter's workshop"\\
    Scene B. "An elegant bathroom featuring a tub, sink, mirror, and decorations"\\
    Generated Question: Which scene is more likely to involve louder noises?\\

    Scene A. "The night sky showcasing the Milky Way"\\
    Scene B. "A bustling city street at midday"\\
    Scene C. "A serene mountain landscape in the morning"\\
    Generated Question: Which scene is different from the other two?\\

    Scene A. "A painting depicting a stormy sea"\\
    Scene B. "A photograph of a calm beach at sunset"\\
    Scene C. "A digital illustration of a bustling space station"\\
    Scene D. "A sculpture of a tranquil garden"\\
    Generated Question: Which scene is most different from the other three?\\
    
    Scene A. "A team of firefighters putting out a blaze in a city"\\
    Scene B. "A family enjoying a picnic in a peaceful park"\\
    Generated Question: Which scene involves a greater sense of danger and urgency?\\
    
    Scene A. "A snowy mountain peak illuminated by the golden light of sunrise"\\
    Scene B. "A tropical beach with crystal-clear water and palm trees swaying in the breeze"\\
    Scene C. "A bustling city park filled with people enjoying outdoor activities"\\
    Scene D. "A vast desert under a blazing sun with sand dunes stretching to the horizon"\\
    Generated Question: Which scene represents a colder and more remote environment? \\ 
    
\end{quote}
We implement a filtering process to exclude questions that focus on textual or difficult to measure qualities. This excludes questions containing terms (and derivatives) such as \textit{\small `word', `text',  `verb', `noun', `describe', `question', `sentence', `detail', `visual', `image', `video', `audio', `sound', `heard', `3d', `point cloud', `caption', `more elements',  `most elements', `more objects',  `more people', `most objects', `more colors', `most colors', `more than one', `similar', `rating', `score'.}

\noindent\textcolor{step3}{\textbf{Step 3: Answer-Explanation Generation}} Building on the captions in the original dataset and the questions refined in \textcolor{step2}{Step 2}, we require the same LLM to answer and explain its reasoning using the following prompt:
\begin{quote}
\small
<s>You are given some scenes described in text as well as a question about them. Each scene is represented by a short caption. Your task is to provide a clear and concise answer that explains the reasoning behind the correct choice. Here are some examples to guide you:\\

    Scene A. "a shattered piece of paper, resembling a broken phone and a flying newspaper"\\
    Scene B. "tourists walking near a catholic church in Mexico on a sunny summer day"\\
    Question: Which scene evokes a sense of chaos and abandonment?\\
    Answer: Scene A. Scene A evokes feelings of chaos and abandonment, contrasting sharply with the joy and vibrancy of Scene B.\\

    Scene A. "Someone is using a rip saw in a carpenter's workshop"\\
    Scene B. "An elegant bathroom featuring a tub, sink, mirror, and decorations"\\
    Question: Which scene is more likely to involve louder noises?\\
    Answer: Scene A. Scene A is characterized by the noise and activity of craftsmanship, whereas Scene B offers a serene and luxurious ambiance for relaxation.\\

    Scene A. "The night sky showcasing the Milky Way"\\
    Scene B. "A bustling city street at midday"\\
    Scene C. "A serene mountain landscape in the morning"\\
    Question: Which scene is different from the other two?\\
    Answer: Scene B. Scene B, with its bustling city life, differs in its dynamic and urban setting from the tranquil and natural settings of Scenes A and C.\\

    Scene A. "A painting depicting a stormy sea"\\
    Scene B. "A photograph of a calm beach at sunset"\\
    Scene C. "A digital illustration of a bustling space station"\\
    Scene D. "A sculpture of a tranquil garden"\\
    Question: Which scene is most different from the other three?\\
    Answer: Scene C. Scene C, a digital illustration of a bustling space station, diverges in its futuristic and technological theme from the natural and serene subjects of the other inputs.\\

    Scene A. "A team of firefighters putting out a blaze in a city"\\
    Scene B. "A family enjoying a picnic in a peaceful park"\\
    Question: Which scene involves a greater sense of danger and urgency?\\
    Answer: Scene A. Scene A, with firefighters responding to a blaze, conveys a strong sense of danger and urgency compared to the calm and leisurely atmosphere of Scene B.\\

    Scene A. "A snowy mountain peak illuminated by the golden light of sunrise"\\
    Scene B. "A tropical beach with crystal-clear water and palm trees swaying in the breeze"\\
    Scene C. "A bustling city park filled with people enjoying outdoor activities"\\
    Scene D. "A vast desert under a blazing sun with sand dunes stretching to the horizon"\\
    Question: Which scene represents a colder and more remote environment?\\
    Answer: Scene A. Scene A, featuring a snowy mountain peak, exemplifies a cold and remote environment in contrast to the other settings, which are warmer or more populated.
\end{quote}
 \textcolor{step4}{\textbf{Step 4: Mixture-of-Models Round-Trip-Consistency (MoM-RTC):}} This step verifies the answers of \textcolor{step3}{Step 3}, via querying multiple models under all possible permutations of the inputs. For clarity, we present a pseudo-algorithm for the MoM-RTC procedure in Algorithm \ref{alg:mom_rtc}. Each of the three LLMs in this procedure is prompted as follows:
 \begin{quote}
    \small
    Select which of the scenes best answers the question. Respond with brevity, and only include your choice in the response.\\
    Question: \{question\}\\
    Choices: \\
    Scene A. \{first modality caption\}\\
    Scene B. \{second modality caption\}\\
    \textit{and so on...}\\
    Answer:
 \end{quote}

\begin{algorithm*}[h]
\small
\caption{Mixture-of-Models Round-Trip-Consistency (MoM-RTC)}
\label{alg:mom_rtc}
\begin{algorithmic}[1]

\Require 
\textit{data}: List of samples, each with \{question, cross-modal info, original\_answer\};\\
\textit{models}: Ensemble of LLMs/classifiers;\\
\textit{permute\_strategy} \(\in \{\text{NONE}, \text{RANDOM}, \text{ALL}\};\)\\
\textit{filtering\_criteria} \(\in \{\text{MF}, \text{UF}, \text{PMF}, \text{PUF}\};\)

\Ensure 
\textit{filtered\_data}: Subset of samples passing the consistency check

\vspace{0.5em}

\Function{generate\_permutations}{sample, strategy}
   \Comment{Returns a list of permuted versions of \textit{sample}}
\EndFunction

\Function{get\_model\_prediction}{model, sample}
   \Comment{Prompts the \textit{model} on the given \textit{sample} and returns a predicted answer}
\EndFunction

\Function{majority\_vote}{answers}
   \Comment{Returns the most frequent answer in \textit{answers}; or handle ties as needed}
\EndFunction

\Function{unanimous\_vote}{answers}
   \Comment{Returns the unique answer if all are identical, else ``no unanimous consensus''}
\EndFunction

\vspace{0.5em}
\State \(\text{filtered\_data} \gets [\,]\)

\ForAll{\(\text{sample} \in \textit{data}\)}
    \State \(\text{original\_answer} \gets \text{sample.original\_answer}\)
    
    \Comment{1. Generate permutations of the sample}
    \State \(\text{permutations} \gets \textsc{generate\_permutations}(\text{sample}, \text{permute\_strategy})\)
    
    \Comment{2. Query each model on each permutation}
    \State \(\text{predictions\_by\_perm} \gets [\,]\)
    \ForAll{\(\text{perm} \in \text{permutations}\)}
        \State \(\text{model\_preds} \gets [\,]\)
        \ForAll{\(\text{model} \in \text{models}\)}
            \State \(\text{pred} \gets \textsc{get\_model\_prediction}(\text{model}, \text{perm})\)
            \State \(\text{model\_preds.append}(\text{pred})\)
        \EndFor
        \State \(\text{predictions\_by\_perm.append}(\text{model\_preds})\)
    \EndFor

    \Comment{3. Check the consistency criteria}
    \If{\(\text{filtering\_criteria} \in \{\text{MF},\;\text{UF}\}\)}
        \Comment{Single (unpermuted) scenario; use the first permutation’s predictions}
        \State \(\text{model\_preds} \gets \text{predictions\_by\_perm}[0]\)
        
        \If{\(\text{filtering\_criteria} = \text{MF}\)}
            \State \(\text{voted\_answer} \gets \textsc{majority\_vote}(\text{model\_preds})\)
            \If{\(\text{voted\_answer} = \text{original\_answer}\)}
                \State \(\text{filtered\_data.append}(\text{sample})\)
            \EndIf
        \ElsIf{\(\text{filtering\_criteria} = \text{UF}\)}
            \State \(\text{unanimous\_answer} \gets \textsc{unanimous\_vote}(\text{model\_preds})\)
            \If{\(\text{unanimous\_answer} \neq \text{"no unanimous consensus"}\)
                  \textbf{and} \(\text{unanimous\_answer} = \text{original\_answer}\)}
                \State \(\text{filtered\_data.append}(\text{sample})\)
            \EndIf
        \EndIf
    \Else
        \Comment{PMF or PUF: multiple permutations}
        \State \(\text{consistent\_across\_all} \gets \text{True}\)
        
        \ForAll{\(\text{model\_preds} \in \text{predictions\_by\_perm}\)}
            \If{\(\text{filtering\_criteria} = \text{PMF}\)}
                \State \(\text{voted\_answer} \gets \textsc{majority\_vote}(\text{model\_preds})\)
                \If{\(\text{voted\_answer} \neq \text{original\_answer}\)}
                    \State \(\text{consistent\_across\_all} \gets \text{False}\)
                    \State \textbf{break}
                \EndIf
            \ElsIf{\(\text{filtering\_criteria} = \text{PUF}\)}
                \State \(\text{unanimous\_answer} \gets \textsc{unanimous\_vote}(\text{model\_preds})\)
                \If{\((\text{unanimous\_answer} = \text{"no unanimous consensus"}) \lor
                       (\text{unanimous\_answer} \neq \text{original\_answer})\)}
                    \State \(\text{consistent\_across\_all} \gets \text{False}\)
                    \State \textbf{break}
                \EndIf
            \EndIf
        \EndFor

        \If{\(\text{consistent\_across\_all}\)}
            \State \(\text{filtered\_data.append}(\text{sample})\)
        \EndIf
    \EndIf

\EndFor
\State \Return \(\text{filtered\_data}\)

\end{algorithmic}
\end{algorithm*}

\section{Category Distribution}
\label{app:category_extraction}
\begin{figure}
    \centering
    \includegraphics[width=\linewidth]{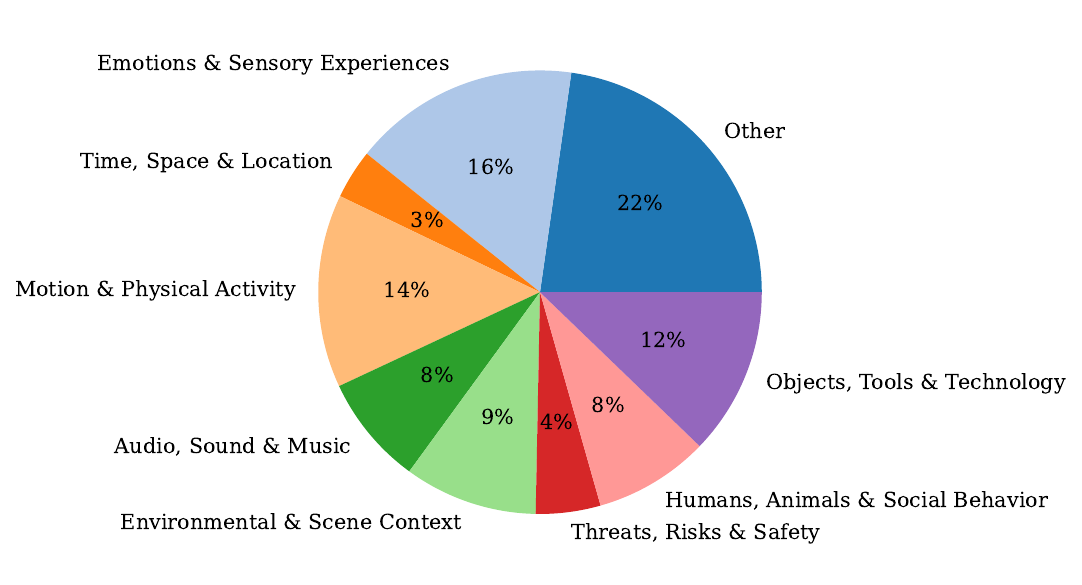}
    \caption{\small Category distribution in human annotated set}
    \label{fig:category_dist}
\end{figure}

To analyze the breadth of the dataset we automatically extract instance categories by employing an LLM which are then grouped based on keyword matching. In particular, we use {\texttt{meta-llama/Llama-3.1-8B-Instruct}} served via VLLM~\cite{kwon2023efficient} and prompt it to predict the topic of each question using the following prompt:
\begin{quote}
\small
You are tasked with categorizing a question that compares or evaluates inputs based on a specific property (e.g., which input is more positive, has more action, etc.).\\

Example Questions and Outputs:\\
Question: "Which input is more positive in tone?"\\
Category: Sentiment Analysis\\
Reasoning: The question explicitly asks about emotional tone, a sentiment-related property.\\

Question: "Which video has more action?"\\
Category: Activity Level\\
Reasoning: The question focuses on the level of dynamism or activity in the input videos.\\

Question: "Which object is larger?"\\
Category: Size Comparison\\
Reasoning: The question compares a specific property, size, between inputs.\\

Question: "Which scene is more likely to involve human presence?"\\
Category: Human Presence\\
Reasoning: The question asks about the likelihood of human presence\\

Question: "Which scene involves more unpredictable or sudden changes?"\\
Category: Dynamic Changes\\
Reasoning: The question asks about the level of unpredictability or sudden changes in the scene.\\

Question: \{question\}\\
Category:
\end{quote}
Figure \ref{fig:category_dist} illustrates the resulting category distribution on the annotated test set. 

\section{Caption Datasets}
\label{app:caption_datasets}
In Table \ref{tab:caption_datasets} we provide details on the captioning datasets used to connect the separate modalities in \texttt{Contra4}.
\begin{table}[ht]
\centering
\fontsize{6}{6}\selectfont
    \centering
    \renewcommand{\arraystretch}{1}
    \begin{tabularx}{\linewidth}{l*{7}{X}}
    \toprule
     Modality    &  Dataset & Train Split & Test Split &  Captions License &  Data License\\\midrule
   \multirow{1}{*}{Image}  & MSCOCO \citep{changpinyo2021conceptual} & Train2017 &Val2017 & CC by 4.0&  CC by 4.0\\\midrule
      \multirow{1}{*}{Video} &  MSRVTT \citep{xu2016msr} & Train & Test & MIT License &MIT License \\\midrule
      3D &  PointLLM \citep{xu2023pointllm} & train & test & ODC-By 1.0 &CC-by-4.0\\\midrule
      \multirow{2}{*}{Audio} & AudioCaps \citep{kim2019audiocaps} & Train &Validation &MIT License& CC by 4.0 \\
       & Clotho \citep{drossos2020clotho} & Development & Evaluation(v1) + Validation(v2) & Non-Commercial&Non-Commercial\\
    \bottomrule
    \end{tabularx}
    \caption{\small Datasets used to generate \texttt{Contra4}}
    \label{tab:caption_datasets}
\end{table}

\section{Additional Dataset Statistics}
\label{app:data_stat}
\begin{figure}[H]
    \centering
    \includegraphics[width=\linewidth]{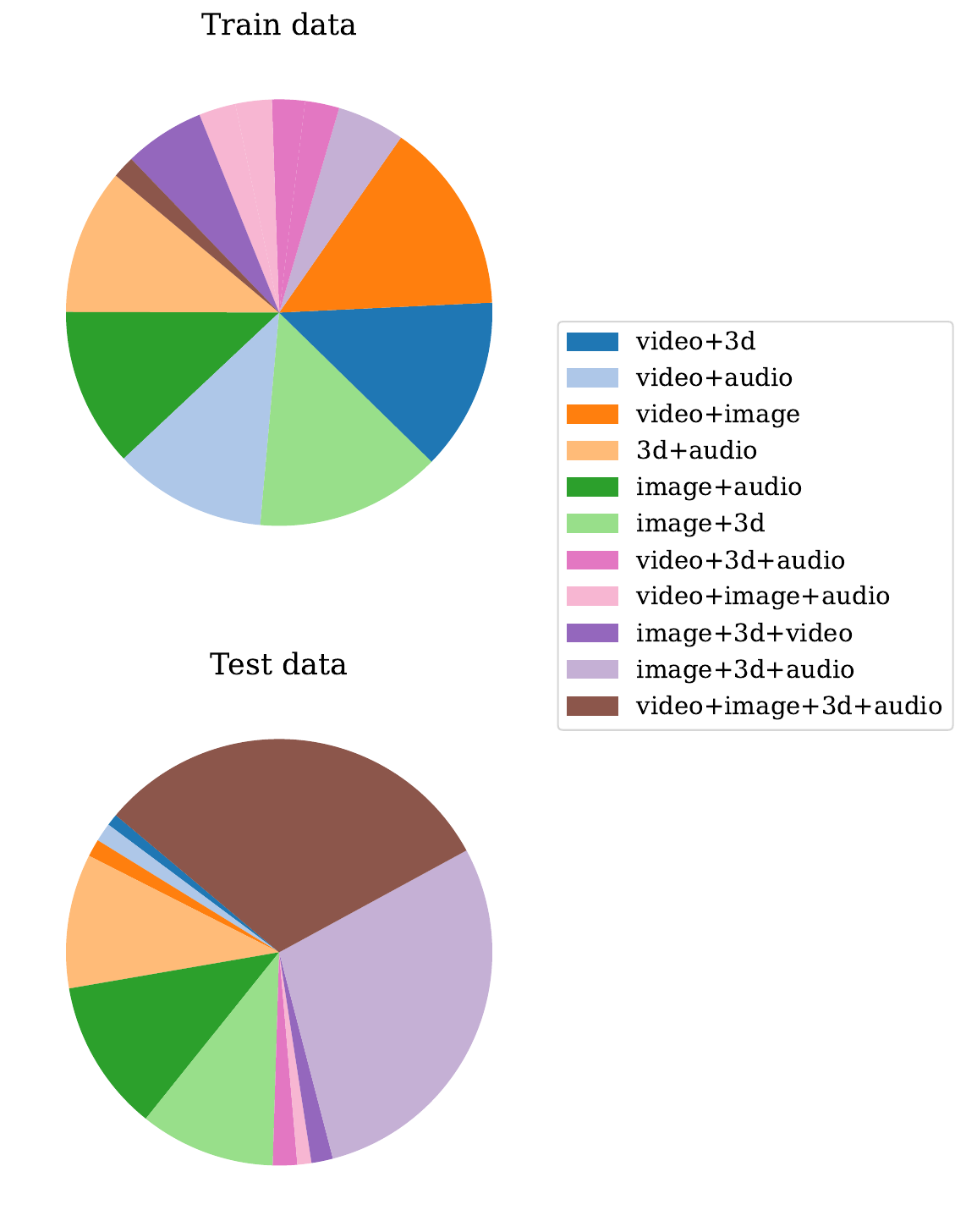}
    \caption{Modality Combination Distribution}
    \label{fig:modality_distribtuion}
\end{figure}

Using the MoM-RTC pipeline, we produce 174k automatically annotated samples for training and release a test set of 2.3k human-annotated examples. Table \ref{tab:human_eval_app} shows a more detailed breakdown on the types of data maintained across different MoM-RTC methods. Fig. \ref{fig:modality_distribtuion} shows the distribution of different modalities in the train and test data. 
\begin{table*}[h]
    \centering
    \fontsize{6}{6}\selectfont
    \renewcommand{\arraystretch}{1}
    \begin{tabularx}{\linewidth}{l *{18}{X}}
    \toprule
     Filter    &  Human Acc. & Recall & N/A & O/A  & GPU~(hrs) & \multicolumn{3}{c}{{2 Modalities}} 
& \multicolumn{3}{c}{{3 Modalities}} 
& \multicolumn{3}{c}{{4 Modalities}}
& \multicolumn{3}{c}{{Aggregated}} \\
     &&&&&& {Rand.} & {Sim.} & {All}
& {Rand.} & {Sim.} & {All}
& {Rand.} & {Sim.} & {All}
& {Rand.} & {Sim.} & {All} \\\midrule
    None$^\ast$ & 39.5 & \textbf{89.1}&13.6&10.2  & 0 & 143k & 145k & 287k & 90k  & 94k  & 183k & 21k  & 23k  & 44k  & 254k & 261k & 515k \\ \midrule    
    MF &65.5&85.7& 7.9 &7.1 &\multirow{2}{*}{40}& 115k & 113k & 227k & 63k  & 62k  & 125k & 13k  & 13k  & 26k  & 190k & 188k & 378k \\
    
    UF   &71.5&79.7& 4.4 &10.9  &&83k  & 80k  & 163k & 40k  & 38k  & 78k  & 8k   & 7k   & 15k  & 130k & 126k & 256k \\
    \midrule
    PMF &72.8& 76.7& 6.6&7.3&\multirow{2}{*}{120}& 102k & 93k  & 196k & 39k  & 34k  & 73k  & 5k   & 4k   & 9k   & 147k & 131k & 278k \\

    PUF & \textbf{83.3}&74.4 &\textbf{0.0}& \textbf{2.5} &  & 69k  & 64k  & 133k & 20k  & 18k  & 38k  & 2k   & 1k   & 3k   & 91k  & 83k  & 174k \\

    \bottomrule
    
    \end{tabularx}
    \caption{\small 
Human Inspection of Different Round-Trip-Consistency Checks on Train Data. 
N/A corresponds to the percentage of wrong examples that are wrong due to lack of applicability to any choice, 
and O/A to the percentage of wrong examples due to the question applying to more than one choice. 
{\scriptsize $^\ast$ some rule based word filtering is applied, see Appendix \ref{app:data_gen}.}
}
    \label{tab:human_eval_app}
\end{table*}

\section{Implementation Details}
\label{app:implementation_details}
All models are served via VLLM~\citep{kwon2023efficient} on 4 A100 40GB GPUs. LLMs are always queried using nucleus sampling with top\_p=0.9. For \textcolor{step2}{Step 2} meta-llama/Llama-3.1-8B-Instruct is queried with temperature equal to 1.05 to encourage diverse questions, and 0.3 for \textcolor{step3}{Step 3} and \textcolor{step4}{Step 4}. All cross-modal LLMs are benchmarked using the default parameter settings in their corresponding repositories and API. For fine-tuning OneLLM we employ LoRA~\cite{hu2021lora} with batch size 8, weight decay 0.02, learning rate 1e-7, and a gradient clipping norm of 2.0 for 10k iterations. {We ensure that modalities are balanced in training via appropriate sampling of all modalities equally. }

\section{Caption Baseline Details}
\label{app:caption_baseline}

 The caption baseline employs OpenGVLab/InternVL2-8B~\cite{chen2024internvl} to generate captions for images, Qwen/Qwen2-VL-7B-Instruct~\cite{Qwen2VL} for videos, Qwen/Qwen2-Audio-7B-Instruct~\citep{Qwen2-Audio} for audio, and X-InstructBLIP~\cite{panagopoulou2023x} for 3D point clouds. With the exception of X-InstructBLIP where we use the official implementation, all other models are queried via VLLM. All models are queried with the default hyperparameters. We use the following prompts: `Describe the [image/audio/3d model]' and `Describe this set of frames. Consider the frames to be a part of the same video.'. Table \ref{tab:pred_cap_scores} shows the captioning performance on each modality for the validation subset of Contra4. The reasoning LLM underlying this baseline is \texttt{meta-llama/Llama-3.1-8B-Instruct}.
\begin{table}[H]
\centering
\small
\begin{tabularx}{.8\linewidth}{l *{4}{c}}
\toprule
       & Image & Video & Audio & 3D \\
\midrule
METEOR & 0.21  & 0.15  &  0.20  & 0.17 \\
\bottomrule
\end{tabularx}
\caption{Predicted caption performance (METEOR)}
\label{tab:pred_cap_scores}
\vspace{-.4cm}
\end{table}

\section{Detailed OneLLM fine-tuning Results}
\label{app:onellm_ft_details}
\begin{figure*}
    \centering
    \includegraphics[width=0.8\linewidth]{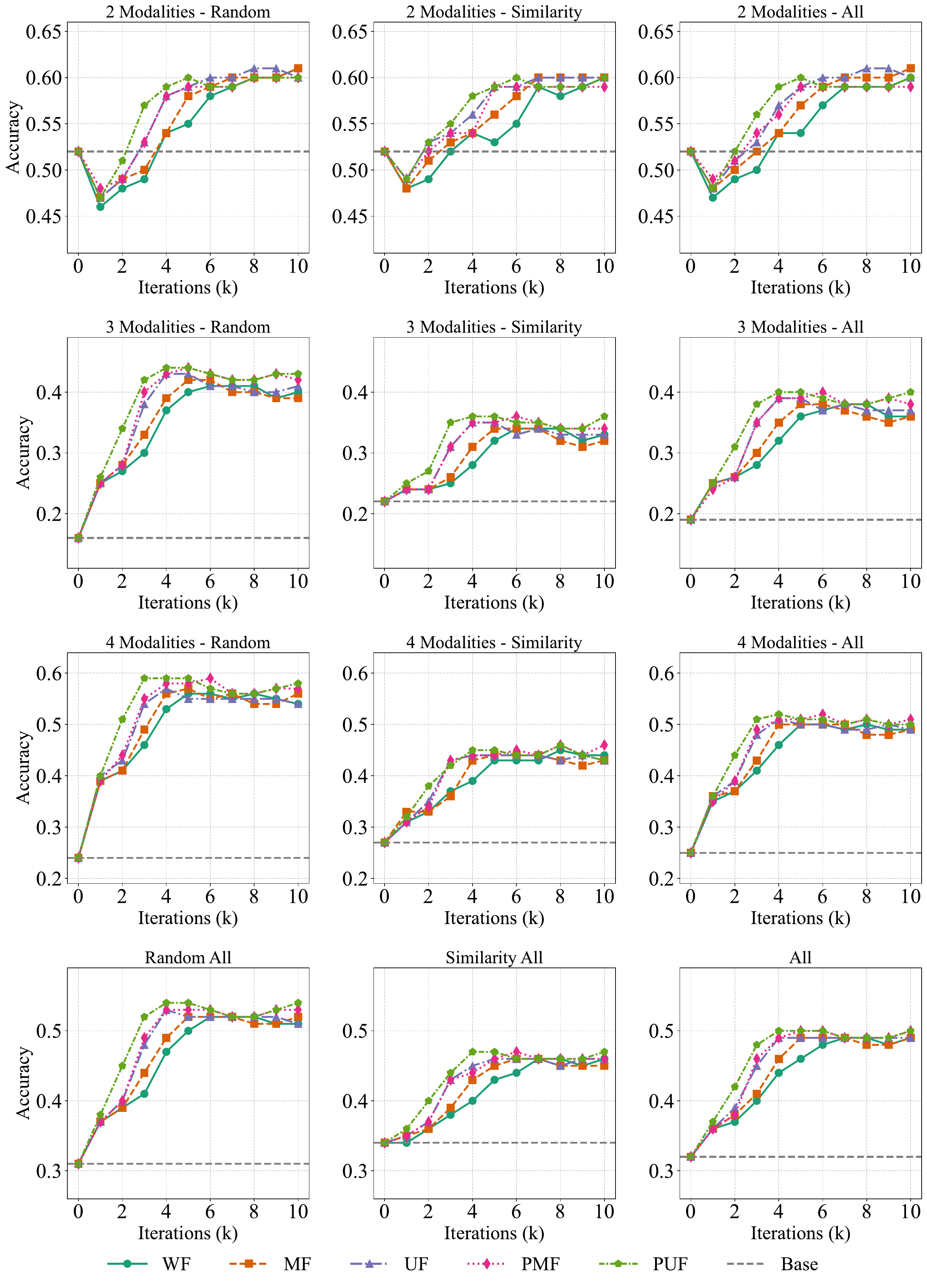}
    \caption{\small Detailed break down of MoM-RTC data effectiveness for fine-tuning OneLLM}
    \label{fig:onellm_ft_full}
\end{figure*}

Figure \ref{fig:onellm_ft_full} shows a breakdown of fine-tuning performance across different question types. We find the trend to be similar across all subsets, with lower performance on examples sampled with high similarity. 

\section{Dataset Examples}
\label{app:qualitative_examples}
Figure \ref{fig:examples} displays data examples from the test split of Contra4 for each of the categories identified in Appendix \ref{app:category_extraction}.
\begin{figure*}
\centering
    \includegraphics[width=\linewidth]{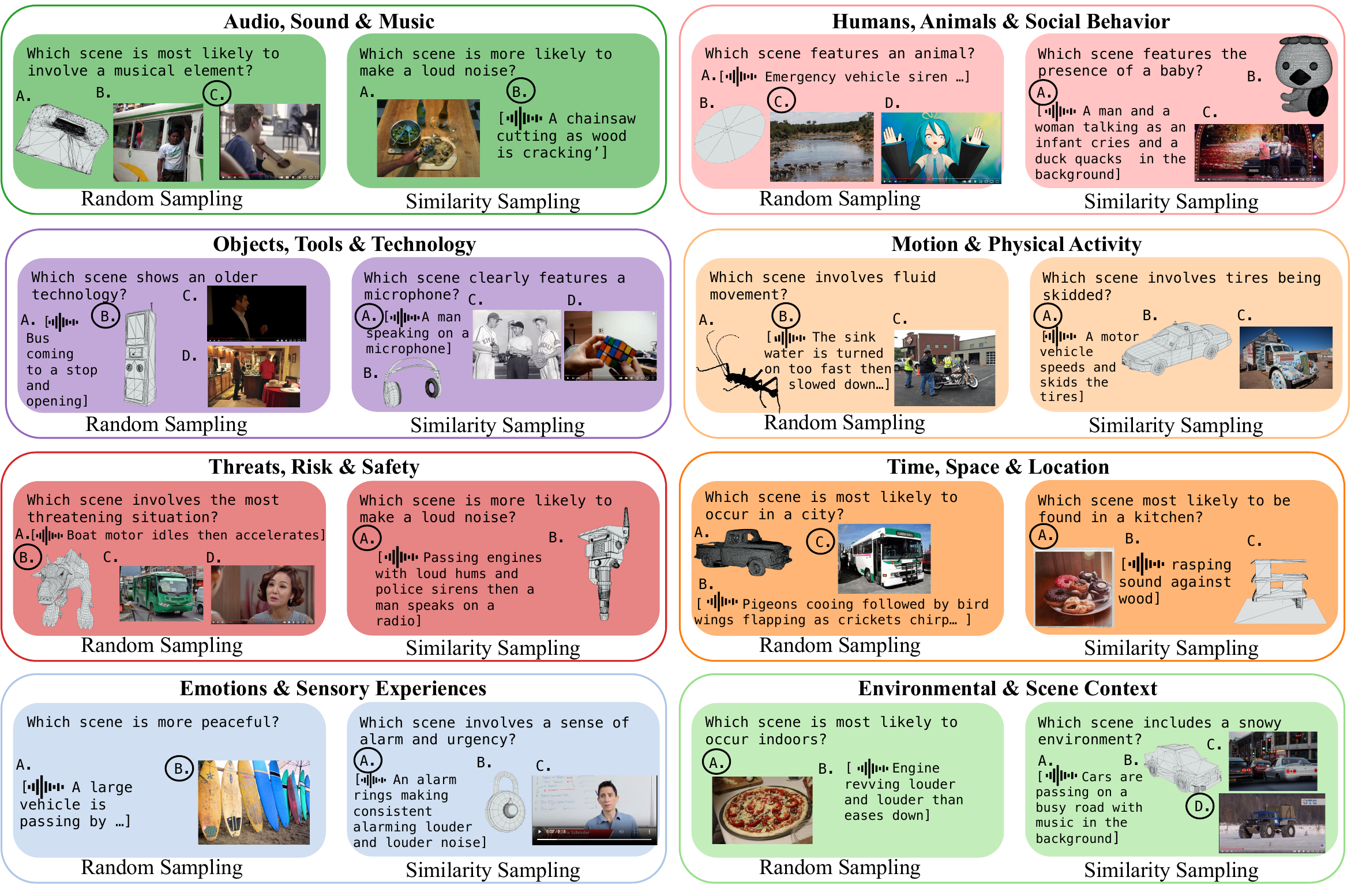}
    \caption{\small Dataset examples for each category.}
    \label{fig:examples}
\end{figure*}

\section{Human Annotation}
\label{app:human_annotation}
\begin{figure*}[h]
    \centering
    \includegraphics[width=.8\linewidth]{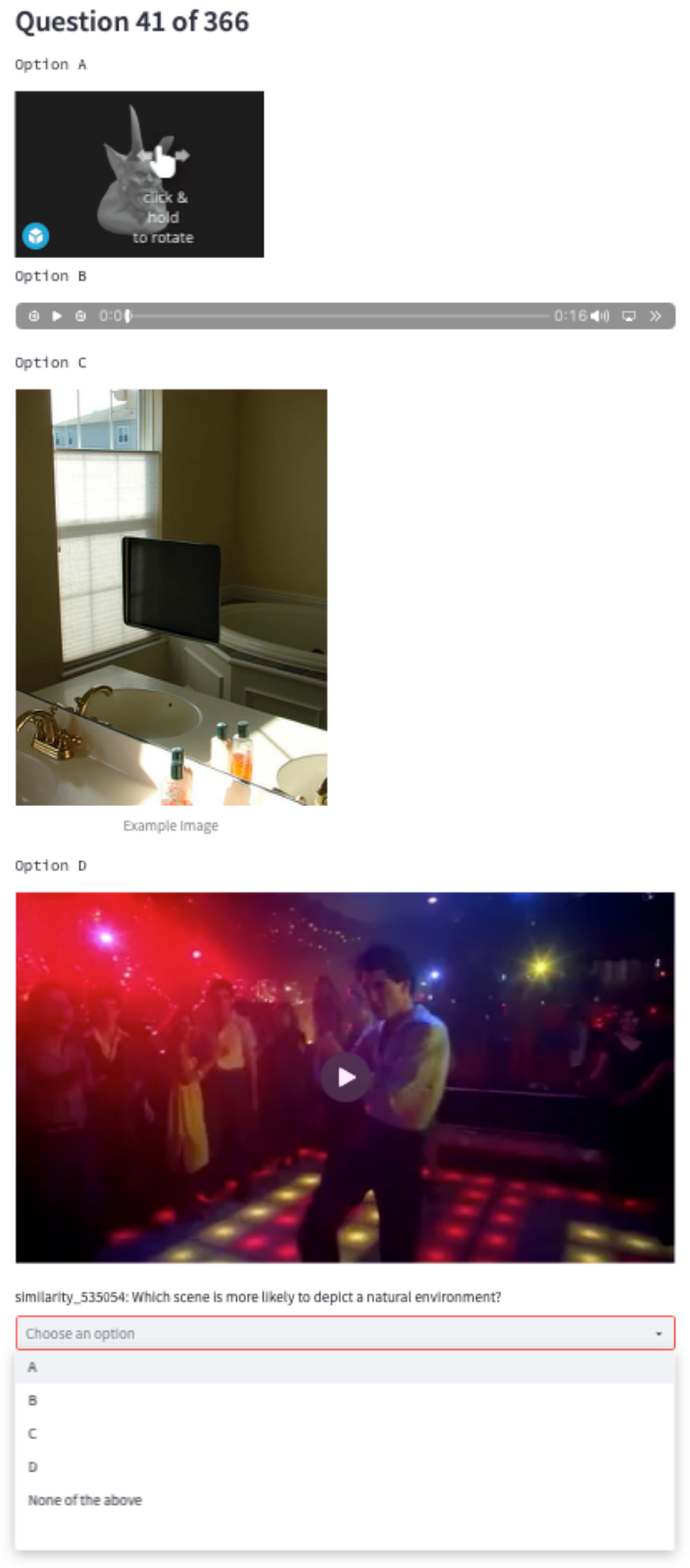}
    \caption{Interface for Human Inspection}
    \label{fig:annotation_ui}
\end{figure*}

In evaluating the effectiveness of mixture-of-models round-trip-consistency for synthetic data generation, we develop a user interface presented in Figure \ref{fig:annotation_ui}. These volunteers were not offered monetary compensation and participated primarily out of academic interest and willingness to contribute to ongoing research as they are all graduate students in computer science in an American university. All annotators provided informed consent and were briefed on the nature of the task prior to participation. For each example, we present the question and the corresponding modality choices, with the option to select `None of the above.' {Note that human annotation was not part of the data generation process itself, as we relied on automatic filtering methods. However, for the test set, all samples were manually reviewed by the authors to ensure clarity, correctness, and overall quality.}

\begin{table}[ht]
\centering
 \fontsize{5}{5}\selectfont
 \renewcommand{\arraystretch}{1}
\begin{tabularx}{\linewidth}{l *{12}{>{\centering\arraybackslash}X}}
\toprule
\multicolumn{13}{c}{\textbf{Correctness}} \\
\midrule
\multicolumn{1}{l}{\textbf{Filter}} 
& \multicolumn{3}{c}{\textbf{2 Modalities}} 
& \multicolumn{3}{c}{\textbf{3 Modalities}} 
& \multicolumn{3}{c}{\textbf{4 Modalities}}
& \multicolumn{3}{c}{\textbf{Aggregated}} \\
\cmidrule(lr){2-4}\cmidrule(lr){5-7}\cmidrule(lr){8-10}\cmidrule(lr){11-13}
& \textbf{Rand.} & \textbf{Sim.} & \textbf{All}
& \textbf{Rand.} & \textbf{Sim.} & \textbf{All}
& \textbf{Rand.} & \textbf{Sim.} & \textbf{All}
& \textbf{Rand.} & \textbf{Sim.} & \textbf{All} \\
\midrule
\textbf{WF}  & 75.0 & 60.0 & 67.5 & 30.0 & 30.0 & 30.0 & 55.0 & 30.0 & 42.5 & 53.3 & 40.0 & 46.7 \\
\textbf{MF}  & 90.0 & 70.0 & 80.0 & 55.0 & 50.0 & 52.5 & 55.0 & 40.0 & 47.5 & 66.7 & 53.3 & 60.0 \\
\textbf{UF}  & 60.0 & 90.0 & 75.0 & 50.0 & 55.0 & 52.5 & 75.0 & 30.0 & 52.5 & 61.7 & 58.3 & 60.0 \\
\textbf{PMF} & 75.0 & 65.0 & 70.0 & 60.0 & 80.0 & 70.0 & 65.0 & 65.0 & 65.0 & 66.7 & 70.0 & 68.3 \\
\textbf{PUF} & 85.0 & 70.0 & 77.5 & 75.0 & 90.0 & 82.5 & 95.0 & 85.0 & 90.0 & 85.0 & 81.7 & 83.3 \\
\midrule

\multicolumn{13}{c}{\textbf{Over-Applies (OA)}} \\
\midrule
\multicolumn{1}{l}{\textbf{Filter}} 
& \multicolumn{3}{c}{\textbf{2 Modalities}} 
& \multicolumn{3}{c}{\textbf{3 Modalities}} 
& \multicolumn{3}{c}{\textbf{4 Modalities}}
& \multicolumn{3}{c}{\textbf{Aggregated}} \\
\cmidrule(lr){2-4}\cmidrule(lr){5-7}\cmidrule(lr){8-10}\cmidrule(lr){11-13}
& \textbf{Rand.} & \textbf{Sim.} & \textbf{All}
& \textbf{Rand.} & \textbf{Sim.} & \textbf{All}
& \textbf{Rand.} & \textbf{Sim.} & \textbf{All}
& \textbf{Rand.} & \textbf{Sim.} & \textbf{All} \\
\midrule
\textbf{WF}  & 5.0 & 10.0 & 7.5 & 30.0 & 20.0 & 25.0 & 10.0 & 35.0 & 22.5 & 15.0 & 21.7 & 18.3 \\
\textbf{MF}  & 0.0 & 20.0 & 10.0 & 5.0 & 15.0 & 10.0 & 20.0 & 45.0 & 32.5 & 8.3 & 26.7 & 17.5 \\
\textbf{UF}  & 0.0 & 5.0 & 2.5 & 30.0 & 15.0 & 22.5 & 0.0 & 30.0 & 15.0 & 10.0 & 16.7 & 13.3 \\
\textbf{PMF} & 15.0 & 20.0 & 17.5 & 20.0 & 5.0 & 12.5 & 15.0 & 15.0 & 15.0 & 16.7 & 13.3 & 15.0 \\
\textbf{PUF} & 0.0 & 0.0 & 0.0 & 5.0 & 15.0 & 10.0 & 10.0 & 5.0 & 7.5 & 5.0 & 6.7 & 5.8 \\
\midrule

\multicolumn{13}{c}{\textbf{None-Applies (NA)}} \\
\midrule
\multicolumn{1}{l}{\textbf{Filter}} 
& \multicolumn{3}{c}{\textbf{2 Modalities}} 
& \multicolumn{3}{c}{\textbf{3 Modalities}} 
& \multicolumn{3}{c}{\textbf{4 Modalities}}
& \multicolumn{3}{c}{\textbf{Aggregated}} \\
\cmidrule(lr){2-4}\cmidrule(lr){5-7}\cmidrule(lr){8-10}\cmidrule(lr){11-13}
& \textbf{Rand.} & \textbf{Sim.} & \textbf{All}
& \textbf{Rand.} & \textbf{Sim.} & \textbf{All}
& \textbf{Rand.} & \textbf{Sim.} & \textbf{All}
& \textbf{Rand.} & \textbf{Sim.} & \textbf{All} \\
\midrule
\textbf{WF}  & 10.0 & 20.0 & 15.0 & 30.0 & 25.0 & 27.5 & 10.0 & 15.0 & 12.5 & 16.7 & 20.0 & 18.3 \\
\textbf{MF}  & 10.0 & 10.0 & 10.0 & 40.0 & 25.0 & 32.5 & 20.0 & 5.0 & 12.5 & 23.3 & 13.3 & 18.3 \\
\textbf{UF}  & 30.0 & 5.0 & 17.5 & 20.0 & 15.0 & 17.5 & 20.0 & 10.0 & 15.0 & 23.3 & 10.0 & 16.7 \\
\textbf{PMF} & 5.0 & 10.0 & 7.5 & 20.0 & 15.0 & 17.5 & 15.0 & 15.0 & 15.0 & 13.3 & 13.3 & 13.3 \\
\textbf{PUF} & 10.0 & 10.0 & 10.0 & 5.0 & 10.0 & 7.5 & 5.0 & 0.0 & 2.5 & 6.7 & 6.7 & 6.7 \\
\bottomrule
\end{tabularx}
\caption{\small Detailed results of human inspection. We report percentages for each of the metrics on the corresponding data subsets.}
\label{tab:human_eval_details}
\end{table}

\end{document}